# Ensemble learning for blending gridded satellite and gauge-measured precipitation data

Georgia Papacharalampous[1,*], Hristos Tyralis[2], Nikolaos Doulamis[3], Anastasios Doulamis[4]

[1] Department of Topography, School of Rural, Surveying and Geoinformatics Engineering, National Technical University of Athens, Iroon Polytechniou 5, 157 80 Zografou, Greece (papacharalampous.georgia@gmail.com, gpapacharalampous@hydro.ntua.gr, https://orcid.org/0000-0001-5446-954X)

[2] Department of Topography, School of Rural, Surveying and Geoinformatics Engineering, National Technical University of Athens, Iroon Polytechniou 5, 157 80 Zografou, Greece (montchrister@gmail.com, hristos@itia.ntua.gr, https://orcid.org/0000-0002-8932-4997)

[3] Department of Topography, School of Rural, Surveying and Geoinformatics Engineering, National Technical University of Athens, Iroon Polytechniou 5, 157 80 Zografou, Greece (ndoulam@cs.ntua.gr, https://orcid.org/0000-0002-4064-8990)

[4] Department of Topography, School of Rural, Surveying and Geoinformatics Engineering, National Technical University of Athens, Iroon Polytechniou 5, 157 80 Zografou, Greece (adoulam@cs.ntua.gr, https://orcid.org/0000-0002-0612-5889)

* Corresponding author

**Abstract**: Regression algorithms are regularly used for improving the accuracy of satellite precipitation products. In this context, satellite precipitation and topography data are the predictor variables, and gauged-measured precipitation data are the dependent variables. Alongside this, it is increasingly recognised in many fields that combinations of algorithms through ensemble learning can lead to substantial predictive performance improvements. Still, a sufficient number of ensemble learners for improving the accuracy of satellite precipitation products and their large-scale comparison are currently missing from the literature. In this study, we work towards filling in this specific gap by proposing 11 new ensemble learners in the field and by extensively comparing them. We apply the ensemble learners to monthly data from the PERSIANN (Precipitation Estimation from Remotely Sensed Information using Artificial Neural Networks) and IMERG (Integrated Multi-satellitE Retrievals for GPM) gridded datasets that span over a 15-year period and over the entire the contiguous United States (CONUS). We also use gauge-measured precipitation data from the Global Historical Climatology Network monthly database, version 2 (GHCNm). The ensemble learners combine the predictions of six machine learning regression algorithms (base learners), namely the multivariate adaptive regression splines (MARS), multivariate adaptive polynomial splines (poly-MARS), random forests (RF), gradient boosting machines (GBM), extreme gradient boosting (XGBoost) and Bayesian regularized neural networks (BRNN), and each of them is based on a different combiner. The combiners include the equal-weight combiner, the median combiner, two best learners and seven variants of a sophisticated stacking method. The latter stacks a regression algorithm on top of the base learners to combine their independent predictions. Its seven variants are defined by seven different regression algorithms, specifically the linear regression (LR) algorithm and the six algorithms also used as base learners. The results suggest that sophisticated stacking performs significantly better than the base learners, especially when applied using the LR algorithm. It also beats the simpler combination methods.

## 1. Introduction

Precipitation data are needed for solving a large variety of water resource engineering problems (e.g., those investigated in Kopsiaftis and Mantoglou 2010, Dogulu et al. 2015, Granata et al. 2016, Széles et al. 2018, Curceac et al. 2020, 2021 and Di Nunno et al. 2022) and can be obtained through either ground-based gauge or satellite networks (Sun et al. 2018). The former networks are known to offer more accurate data, while the latter are in general more spatially dense because of their lower cost (Mega et al. 2019, Salmani-Dehaghi and Samani 2021, Li et al. 2022, Tang et al. 2022). In this view, the widely adopted strategy of blending satellite and gauge-measured precipitation data for forming new precipitation products with higher accuracy than the purely satellite ones and higher spatial density than the gauge-measured ones is reasonable. Besides, the importance of the more general problem of merging gridded satellite and gauge-measured data for earth observation is well-recognised, as proven by both the number and the diversity of the relevant studies (see, e.g., Gohin and Langlois 1993, Journée and Bertrand 2010 and Peng et al. 2021). Moreover, as noted in Tyralis et al. (2023), even products that already rely on both gridded satellite and gauge-measured data could be further improved in terms of their accuracy by using gauge-measured data in post-processing frameworks. Therefore, in what follows, data from such products and data from purely satellite products will not be explicitly distinguished, as the focus herein will be on their merging with gauge-measured data.

Machine learning regression algorithms (Hastie et al. 2009, James et al. 2013 and Efron and Hastie 2016) are regularly used for blending multiple precipitation datasets with differences in terms of spatial density and accuracy (see the relevant reviews by Hu et al. 2019 and Abdollahipour et al. 2022, as well as examples of such studies in Tao et al. 2016, Baez-Villanueva et al. 2020, Chen et al. 2020, Chen et al. 2021, Shen and Yong 2021, Zhang et al. 2021, Chen et al. 2022, Lin et al. 2022, Papacharalampous et al. 2023a and Tyralis et al. 2023). In such spatial downscaling (and, more generally, spatial interpolation) settings, satellite precipitation data and topography factors are the predictor variables, and gauged-measured precipitation data are the dependent variables. Alongside these, it is increasingly recognized in many hydrological disciplines that combining individual algorithms through ensemble learning, a concept from the machine learning field that is alternatively referred to as "forecast combinations" in the forecasting field, can lead to considerable improvements in terms of predictive performance (see, e.g., the relevant discussions in the review by Papacharalampous and Tyralis 2022). Detailed reviews on ensemble learning methods can be found in Sagi and Rokach (2018) and Wang et al. (2022).

Aside from the applications of a few individual (ready-made) algorithms that rely on the construction of ensemble learning, such as the random forecast (RF; Breiman 2001) and boosting (Friedman 2001, Chen and Guestrin 2016) ones, this concept has not spread in the literature that fuses gridded satellite products and ground-based precipitation datasets. Indeed, ensemble learning methods, such as the simple ones appearing in Bates and Granger ([1969](#)), Petropoulos and Svetunkov ([2020](#)), Papacharalampous et al. ([2019](#)) and Papacharalampous and Tyralis ([2020](#)), and the more sophisticated ones appearing in Wolpert ([1992](#)), Divina et al. ([2018](#)), Yao et al. ([2018](#)), Alobaidi et al. ([2019](#)), Cui et al. ([2021](#)), Tyralis et al. ([2021](#)), Wu et al. ([2021](#)) and Hwangbo et al. ([2022](#)), have not been proposed and extensively compared with each other in this literature. An exception to this can be found in the study by Zandi et al. ([2022](#)), who introduced a sophisticated ensemble learning algorithm. This algorithm adopts a stacking strategy (Wolpert [1992](#)) and the least absolute shrinkage and selection operator (LASSO; Tibshirani [1996](#)) algorithm to combine the independent predictions of three individual algorithms. The latter are the multilayer perceptron neural networks, support vector machines (Cortes and Vapnik [1995](#)) and RF. Still, many more ensemble learners and large-scale comparisons of them using gauge-measured data and multiple satellite products for an extensive time period and a large geographical region are needed in the field for reaching higher accuracy levels. In this study, we work towards filling in this specific gap. More precisely, our objectives are to propose a large number of new ensemble learners in the field, as well as to extensively compare them.

The remainder of this paper is structured as follows: [Section 2](#) lists and describes the ensemble learning methods introduced and compared in the field by this work. [Section 3](#) presents the data and the framework that facilitated the comparison. [Section 4](#) presents the results. [Section 5](#) discusses the importance of the findings in light of the literature and provides ideas for exploiting the proposed framework for future research. The manuscript concludes with [Section 6](#). For the sake of completeness, [Appendix A](#) lists and briefly describes the machine and statistical learning regression algorithms on which the comparison relied, and [Appendix B](#) provides statistical software information, ensuring the reproducibility of the work and its methods.

## 2. Ensemble learners and combiners

### 2.1 Ensemble learners

In this work, 11 ensemble learners (see [Figure 1](#)) were compared for improving the accuracy of satellite precipitation datasets. Each of these learners combined predictions that were delivered independently by six regression algorithms. The latter are referred to as "base learners" throughout the work and include the multivariate adaptive regression splines (MARS), multivariate adaptive polynomial splines (poly-MARS), RF, gradient boosting machines (GBM), extreme gradient boosting (XGBoost) and Bayesian regularized neural networks (BRNN), which are briefly described in Appendix A. These learners are regularly applied in hydrology (see, e.g., the applications by Rezaali et al. 2021, Granata et al. 2022, Quilty et al. 2022 and Kopsiaftis 2023). Also notably, they are largely diverse with each other and exhibit considerably good performances for the task of interest according to Papacharalampous et al. (2023a). Therefore, their use as base learners is reasonable. While the base learners were the same for all the 11 ensemble learners, the procedures followed for the combination of the independent predictions were different. These procedures are known as "combiners" in the literature of ensemble learning and, in this work, they included the mean combiner (else referred to as the "equal-weight combiner" in the literature and throughout this work; see Section 2.2), the median combiner (see Section 2.3), two best learners (see Section 2.4) and seven variants of a sophisticated stacked generalization method (see Section 2.5).

## 2.2 Mean combiner

The mean combiner performs simple averaging of the independent predictions delivered by the base learners. Thus, it does not require training.

## 2.3 Median combiner

The median combiner computes the median of the independent predictions delivered by the base learners. Similar to the mean combiner, it does not require training.

## 2.4 Best learners

The best learners are ensemble learners that select one base learner, specifically the one that scores the best in a predictive performance comparison on a training dataset, and then adopt the predictions of the selected base learner as their own for all the new problems that are set. In this work, we applied two best learners, one that bases the identification of the best base learner on the mean squared error (MSE) and another that bases this identification on the median squared error (MdSE).

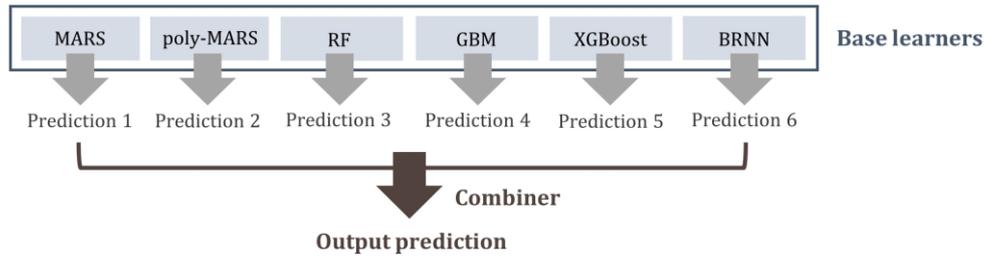

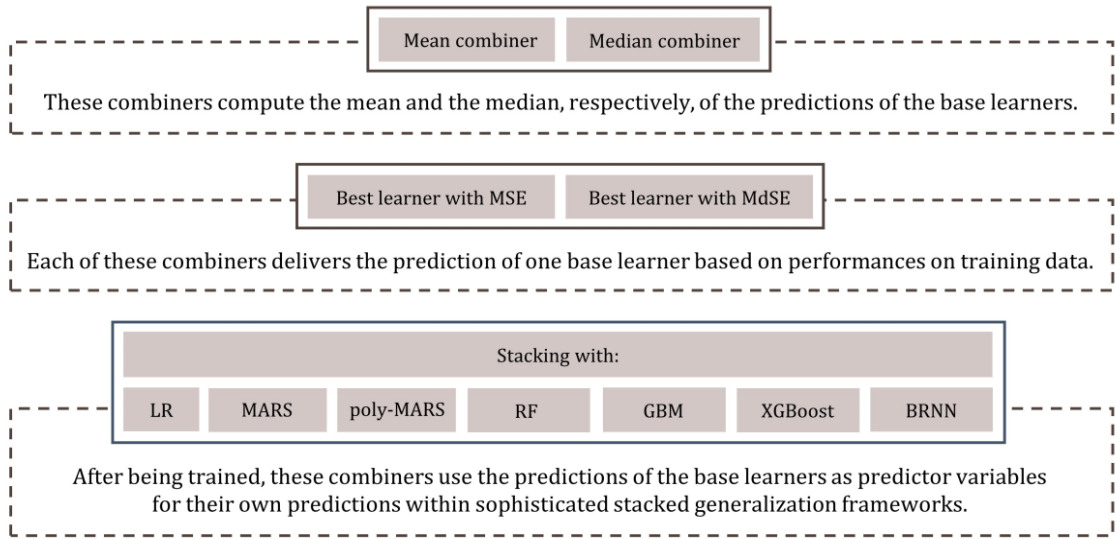

Figure 1. Plain language summary of the ensemble learners of this study. Linear regression, multivariate adaptive regression splines, multivariate adaptive polynomial splines, random forests, gradient boosting machines, extreme gradient boosting, Bayesian regularized neural networks, mean squared error and median squared error are referred to under their abbreviations (i.e., LR, MARS, poly-MARS, RF, GBM, XGBoost, BRNN, MSE and MdSE, respectively).

## 2.5 Stacking of regression algorithms

The strategy of stacking a regression algorithm on top of others facilitates more sophisticated combinations than the mean combiner, median combiner and best learners. In this ensemble learning strategy, the base learners produce predictions independently of each other for a dataset. This dataset plays the role of a training dataset for a selected regression algorithm, which is hereafter referred to as the "meta-learner". More precisely, the predictions of the base learners for this dataset are used together with their corresponding true values for training the meta-learner. The trained meta-learner is then given predictions of the base learners for new problems and delivers its own predictions based on them. For facilitating an extensive comparison between ensemble learners, as well as between combiners, seven variants of this stacking methodology were applied in this work, with each of these variants relying on a different meta-learner. The meta-learners were the linear regression (LR) algorithm and the six algorithms also used as base learners (see their brief descriptions in Appendix A).

## 3. Data and application

### 3.1 Data

We applied the methods to open data. The gauge-measured precipitation data are described in Section 3.1.1, the satellite precipitation data are described in Section 3.1.2 and the elevation data are described in Section 3.1.3.

*3.1.1 Gauge-measured precipitation data*

We used data from the Global Historical Climatology Network monthly database, version 2 (GHCNm; Peterson and Vose 1997). From this database, we extracted total monthly precipitation data in the time period 2001–2015 from 1 421 stations in the contiguous United States (CONUS). The locations of these stations are depicted in Figure 2. The data were sourced from the National Oceanic and Atmospheric Administration (NOAA) data repository (https://www.ncei.noaa.gov/pub/data/ghcn/v2; accessed on 2022-09-24).

*3.1.2 Satellite precipitation data*

Satellite precipitation data from the CONUS and for the same time period (i.e., 2001–2015) were also used. We sourced these data from two repositories. The first of them is the current operational PERSIANN (Precipitation Estimation from Remotely Sensed Information using Artificial Neural Networks) database (Hsu et al. 1997, Nguyen et al. 2018, 2019), which was created by the Centre for Hydrometeorology and Remote Sensing (CHRS) at the University of California, Irvine (UCI). The grid extracted from this database has a spatial resolution of 0.25 degree x 0.25 degree and covers the region of interest entirely (see Figure 3a). In particular, the PERSIANN precipitation data are daily and can be extracted from the Center for Hydrometeorology and Remote Sensing (CHRS) repository (https://chrsdata.eng.uci.edu; accessed on 2022-03-07). We transformed the daily data into total monthly precipitation data to be consistent with the gauge-measured data.

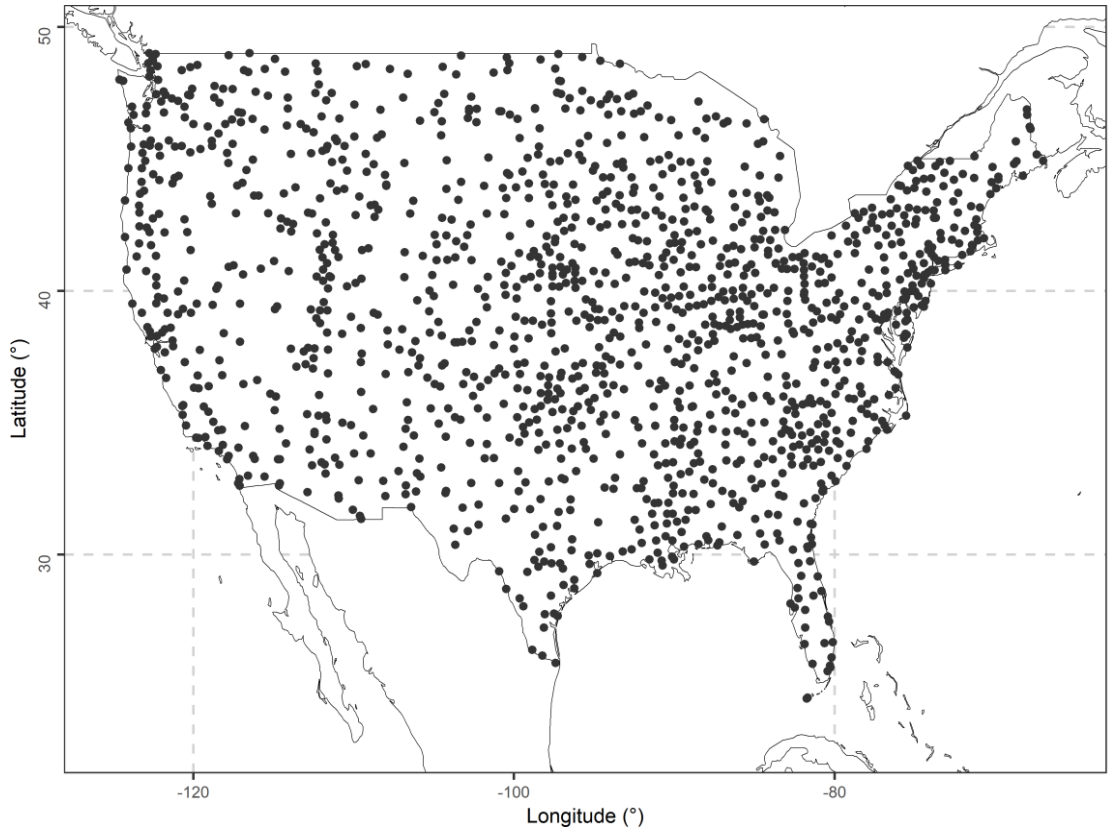

Figure 2. Locations of the ground-based precipitation stations.

The second gridded satellite precipitation dataset used in this work is the GPM IMERG (Integrated Multi-satellitE Retrievals) late Precipitation L3 1 day 0.1 degree x 0.1 degree V06 dataset, which was developed by the NASA (National Aeronautics and Space Administration) Goddard Earth Sciences (GES) Data and Information Services Center (DISC) (Huffman et al. 2019) and is available at the repository of NASA Earth Data (https://doi.org/10.5067/GPM/IMERGDL/DAY/06; assessed on 2022-12-10). After its extraction, the original GPM IMERG daily data were transformed to the CMORPH0.25 grid with a spatial resolution of 0.25 degree x 0.25 degree (see Figure 3b) with a bilinear interpolation procedure. The IMERG data covers the region of interest entirely. Similar to the PERSIANN data, we transformed the daily data into total monthly precipitation data to be consistent with the gauge-based measurements.

Notably, both the PERSIANN and IMERG families of gridded satellite precipitation products are extensively used and examined in the literature (see the relevant examples of studies by Wang et al. 2017, Akbari Asanjan et al. 2018, Nguyen et al. 2018, Jiang et al. 2018, Tan and Santo 2018, Moazami et al. 2022, Pradhan et al. 2022 and Salehi et al. 2022). Therefore, we believe that the above-outlined data can effectively support the comparison of this study.

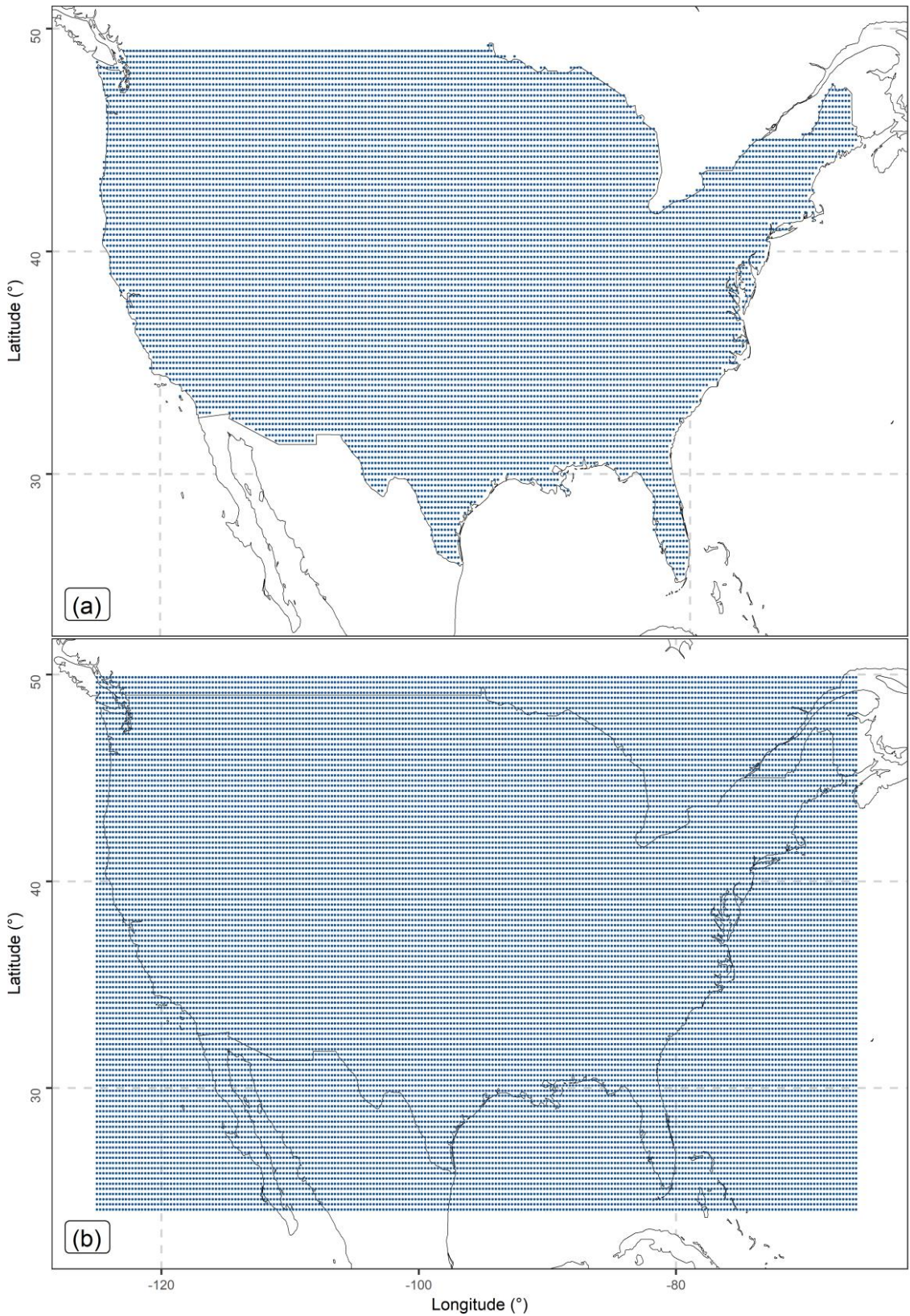

Figure 3. Geographical locations of the (a) PERSIANN and (b) IMERG grid points.

*3.1.3 Elevation data*

As it is highlighted by Xiong et al. (2022), elevation is a useful predictor variable for many hydrological processes. Thus, its value was estimated at all the gauged geographical locations presented in Figure 2 by extracting point elevation data from the Amazon Web Services (AWS) Terrain Tiles (https://registry.opendata.aws/terrain-tiles; accessed on 2022-09-25).

## 3.2 Regression settings and validation procedure

The dependent variable is the gauge-measured total monthly precipitation at a point of interest. According to procedures proposed in Papacharalampous et al. (2023a, b) and Tyralis et al. (2023), we formed the regression settings by finding, separately for the PERSIANN and IMERG grids (see Figure 3a and b, respectively), the four closest grid points to each of the geographical locations of the precipitation ground-based stations (see Figure 2) and by computing the respective distances $d_i$, $i$ = 1, 2, 3 and 4 (in meters). We also indexed these four grid points $S_i$, $i$ = 1, 2, 3 and 4 according to their distance from the stations, where $d_1 < d_2 < d_3 < d_4$ (see Figure 4). Hereinafter, the distances $d_i$, $i$ = 1, 2, 3 and 4 will be called "PERSIANN distances 1, 2, 3 and 4" or "IMERG distances 1, 2, 3 and 4" and the total monthly precipitation values at the grid points 1, 2, 3 and 4 will be called "PERSIANN values 1, 2, 3 and 4" or "IMERG values 1, 2, 3 and 4". We formulated and used three sets of predictor variables, which will be simply called "predictor sets 1, 2 and 3" throughout this work and correspond to three different regression settings. These settings facilitated three different comparisons between the learners. Predictor sets 1, 2 and 3 are, respectively, defined as follows: {PERSIANN values 1, 2, 3 and 4, PERSIANN distances 1, 2, 3 and 4, station elevation}, {IMERG values 1, 2, 3 and 4, IMERG distances 1, 2, 3 and 4, station elevation}, and {PERSIANN values 1, 2, 3 and 4, IMERG values 1, 2, 3 and 4, PERSIANN distances 1, 2, 3 and 4, IMERG distances 1, 2, 3 and 4, station elevation}.

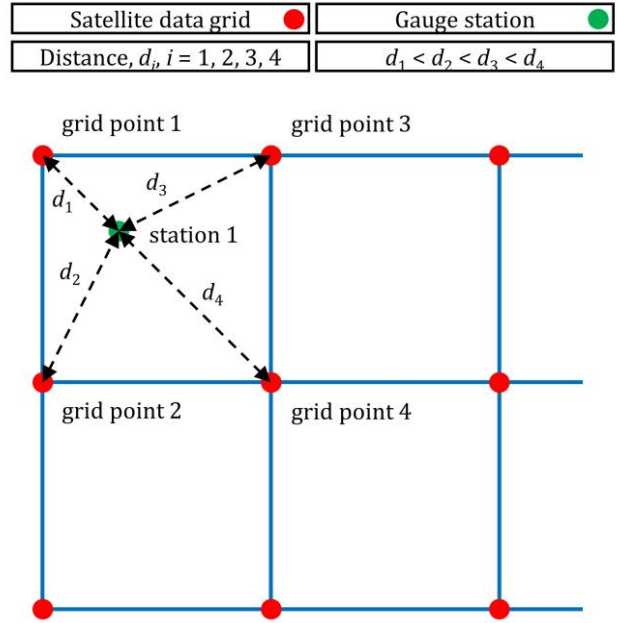

Figure 4. Formulation of the spatial interpolation problem. The IMERG or PERSIANN precipitation data are measured at the grid points 1, 2, 3 and 4. The distances of these grid points from station 1 (i.e., the ground-based station that is surrounded by the same grid points) are denoted with $d_i$, $i$ = 1, 2, 3 and 4.

Each regression setting consisted of 91 623 samples and was divided into three datasets of equal length (hereafter referred to as "datasets 1, 2 and 3"). The division was made randomly; therefore, the various different hydroclimatic and topographic conditions are represented equally well in all the three datasets of each regression setting. The same holds for possible biases in the gridded satellite products. We first trained the six base learners on dataset 1 and applied them to obtain predictions for dataset 2. These predictions were then used by the best learners and the stacking variants (as described in Sections 2.4 and 2.5, respectively) for selecting a best base learner and training the meta-learner, respectively. Then, we trained the base learners on {dataset 1, dataset 2} and applied them to obtain predictions for dataset 3. These latter predictions were used by the mean combiner, the median combiner, the best learners and the stacking variants (as described in Sections 2.2–2.5, respectively) for forming their predictions for dataset 3. They were additionally used for assessing the base learners, as these are also used as benchmarks for the ensemble learners.

3.3  Predictive performance comparison

To compare the 17 learners (i.e., the six base learners and the 11 ensemble learners), we first computed the squared error scoring function, which is defined as

$$S(x, y) := (x - y)^2, \tag{1}$$

where $y$ is the materialization of the spatial process and $x$ is the respective prediction. The squared error scoring function is consistent for the mean functional of the predictive probability distribution (Gneiting 2011). The performance criterion takes the form of MSE, which is defined as

$$\text{MSE} := (1/n) \sum_{i=1}^{n} S(x_i, y_i), \tag{2}$$

where $x_i$ and $y_i$, $i \in \{1, ..., n\}$ are the predictions and materialization of the process, respectively. MSE is computed separately for each set {learner, predictor set}.

Subsequently, skill scores were computed for each set {learner, predictor set}, according to Equation (3). More precisely, type-1 skill scores were computed by using as the reference case (benchmark) the MARS algorithm when run with the same predictor set as the modelling approach to which the relative score referred. On the other hand, type-2 skill scores were computed by considering the set {MARS, predictor set 1} as the reference case (benchmark) for all the sets {learner, predictor set}. Type-1 and type-2 skill scores were used to rank the learners from the best-performing to the worst-performing. The results of the predictive performance comparison are presented in Section 4.1.

$$\text{RS}_{\{\text{learner, predictor set}\}} := 100 \, (1 - \text{MSE}_{\{\text{learner, predictor set}\}} / \text{MSE}_{\text{benchmark}}) \tag{3}$$

## 3.4 Additional investigations

Aside from comparing the 11 ensemble learners of this work in terms of their predictive performance in the context of interest, we also recorded and compared the computational time required for their application. The results of this comparison are available in Section 4.2. Moreover, we applied explainable machine learning (see the relevant reviews by Roscher et al. 2020, Belle and Papantonis 2021, Linardatos et al. 2021) to rank the base learners based on their contribution in ensemble learning using stacking. More precisely, we implemented procedures of the RF algorithm to compute the permutation importance score for the predictions produced by the base learners for dataset 2 (on which the training of the regression algorithm that combines the base learners is made; see Section 3.2) and ranked them based on this score. We additionally implemented procedures of the XGBoost algorithm to compute the gain score for the predictions produced by the base learners for dataset 2 and then determined the corresponding rankings. The results of these two comparisons of the base learners are available in Section 4.3. Lastly, we implemented the above-outlined explainable machine learning procedures to compare the total of the predictor variables (see the predictor set 3 in Section 3.2) in terms of their relevance in solving the regression problem investigated. The results of the comparisons of the predictor variables are available in Section 4.4.

## 4. Results

### 4.1 Predictive performance

Figure 5 presents a comparison based on MSE between the learners. According to this comparison, the six base learners (i.e., MARS, polyMARS, RF, GBM, XGBoost and BRNN) exhibit notable differences in their performance, with RF and XGBoost being the most accurate and MARS being the least accurate for all three predictor sets investigated. Moreover, ensemble learning using the mean combiner gave more accurate predictions than ensemble learning using the median combiner. The latter ensemble learning approach produced less accurate predictions than both RF and XGBoost (with a single exception concerning predictor set 2). Ensemble learning using the mean combiner was also more accurate than all the base learners for predictor sets 2 and 3. The same, however, was not the case for predictor set 1. Indeed, for this predictor set, both the simple combination approaches investigated in this work were found to be less accurate than both RF and XGBoost.

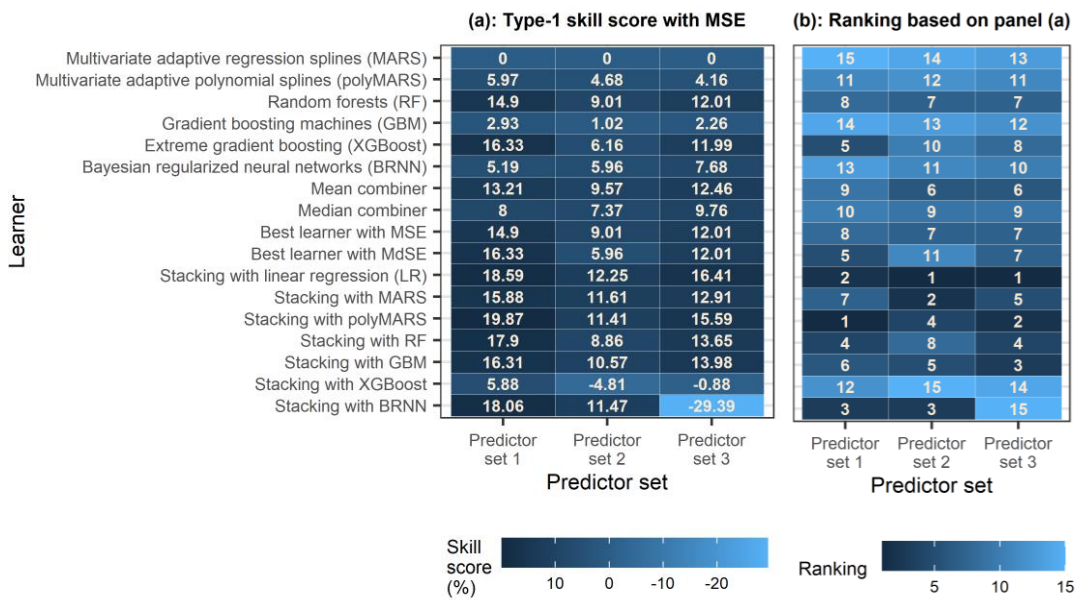

Figure 5. Heatmaps of the: (a) type-1 skill score using the MSE metric; and (b) ranking of each learner according to the MSE metric. The darker the colour, the better the predictions.

Furthermore, the best learners performed exactly the same with RF, XGBoost or BRNN

(as these three base learners were identified as the best on dataset 2) and, thus, better than the median combiner (with one single exception; see above) and worse than the mean combiner for predictor sets 2 and 3, but not for predictor set 1. The best among all the ensemble learning methods that use regression algorithms as meta-learners were stacking with LR and stacking with polyMARS, while the worst was stacking with XGBoost (probably due to the importance of parametrization in the implementation of this latter regression algorithm) for predictor sets 1 and 2, and stacking with BRNN for predictor set 3 due to outliers. The remaining ensemble learning methods using regression algorithms as meta-learners were more accurate than or at least as accurate as the base learners, the simple combination methods and the best learners with a few exceptions concerning stacking with MARS and stacking with GBM for predictor set 1, and stacking with RF for predictor set 2.

Figure 6 additionally allows us to compare the three predictor sets investigated in this work in terms of their usefulness, as they are based on the type-2 skill score. Overall, we can state that predictor set 2 (for which the type-2 skill scores ranged from 17.27% to 30.73% and the respective rankings ranged from 12 to 33) is more useful than predictor set 1 (for which the type-2 skill scores ranged from 0 to 19.87% and the respective rankings ranged from 29 to 45). We can also state that combining predictor sets 1 and 2 into predictor set 3 leads to substantial performance improvements for both the base learners and the ensemble learning methods. Indeed, for predictor set 3, the type-2 skill scores ranged between 27.32% and 39.77%, and the respective rankings ranged between 1 and 21, with a single exception concerning stacking with BRNN (for which the type-2 skill score was 6.77 and the respective ranking was 40).

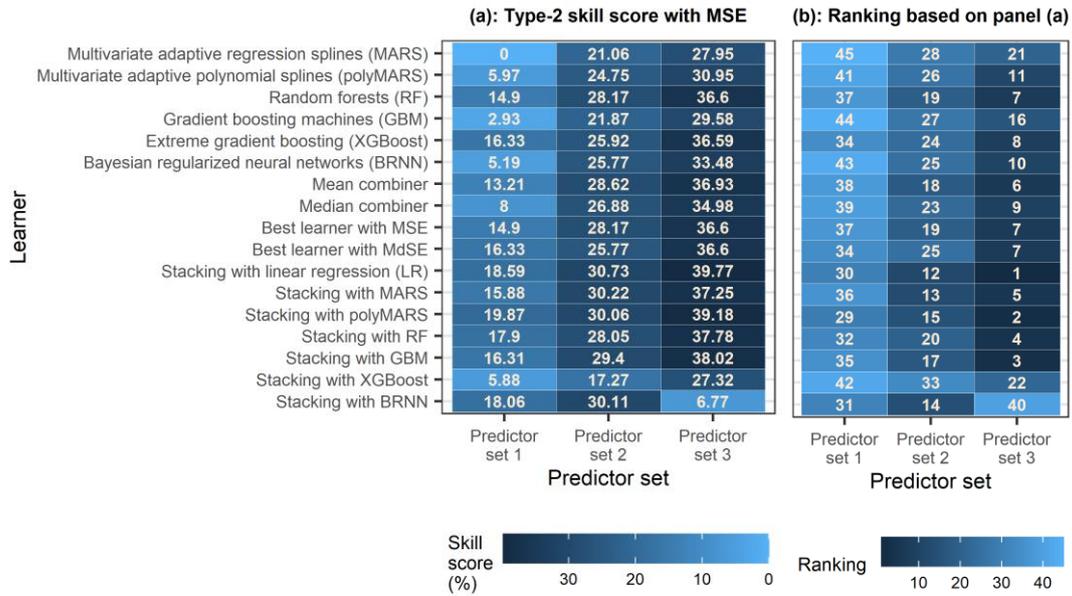

Figure 6. Heatmaps of the: (a) type-2 skill score using the MSE metric; and (b) ranking of each set {learner, predictor set} according to the MSE metric. The darker the colour, the better the predictions.

## 4.2 Computational time

Figure 7 facilitates comparisons of the learners in terms of computational time. Notably, BRNN is by far the most computationally expensive algorithm used as both a base learner and a meta-learner in this work and, probably due to its Bayesian regularization procedure, its computational requirements can differ a lot from dataset to dataset even when the number of predictors is the same (see, e.g., the measured time for predictor sets 1 and 2) and especially when this number differs considerably (see, e.g., the measured time for predictor set 3 compared to the measured time for predictor sets 1 and 2). Simple combination methods simply require the sum of the computational time required by each of the base learners increased by a small amount, which is practically the time required for computing the mean or the median of the predictions for each data sample. The same, however, does not apply to the remaining ensemble learning methods, which are more computationally expensive, because they require two or three different trainings. As the time required to train the algorithms on the predictions of the base learners is much less than the time required for training the base learners, these methods have comparable computational requirements, with the ensemble learning method using BRNN for combining the predictions of the base learners being the most computationally expensive overall.

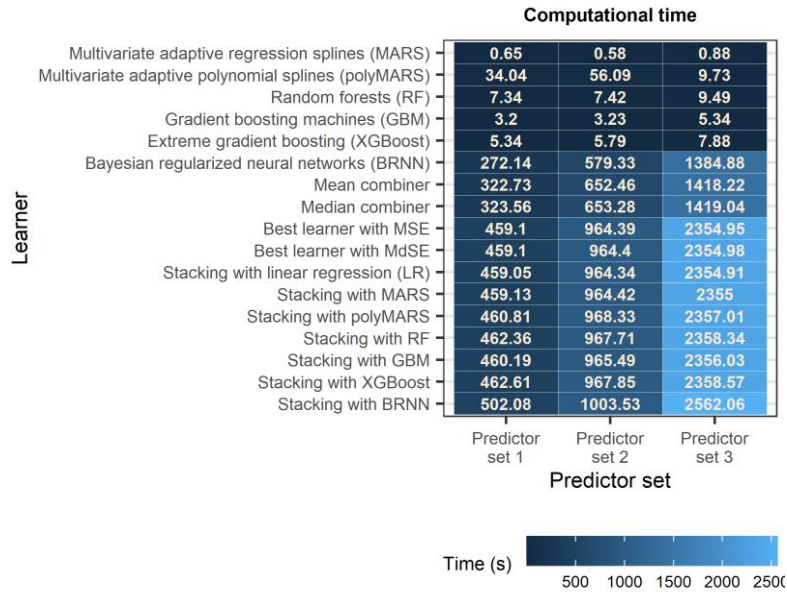

Figure 7. Computational time on an Intel(R) Core(TM) i9-10900 CPU @ 2.80GHz 2.81 GHz with 128 GB RAM. The darker the colour, the lower the computational time.

## 4.3 Contribution of base learners

Figure 8 presents the comparison of the base learners in terms of their contribution to ensemble learning using RF as the meta-learner. The rankings of the base learners based on permutation importance are not exactly the same across the predictor sets. Still, there are notable similarities, with XGBoost and RF being the most important and the second most important base learners, respectively, for all three predictor sets investigated in this study. This latter finding seems reasonable in light of the better predictive performance of XGBoost and RF compared to the remaining base learners (see Figure 5). In addition, Figure 9 presents the comparison of the base learners in terms of their contribution to ensemble learning using XGBoost as the meta-learner. Based on the gain scores (which are computed by taking each feature's contribution to each tree in the model), the three most important base learners for generating a prediction are RF, BRNN and XGBoost. This result is again reasonable, as BRNN also exhibits a better predictive performance than MARS, polyMARS and GBM (see Figure 5), with a single exception concerning predictor set 1, for which polyMARS performed better.

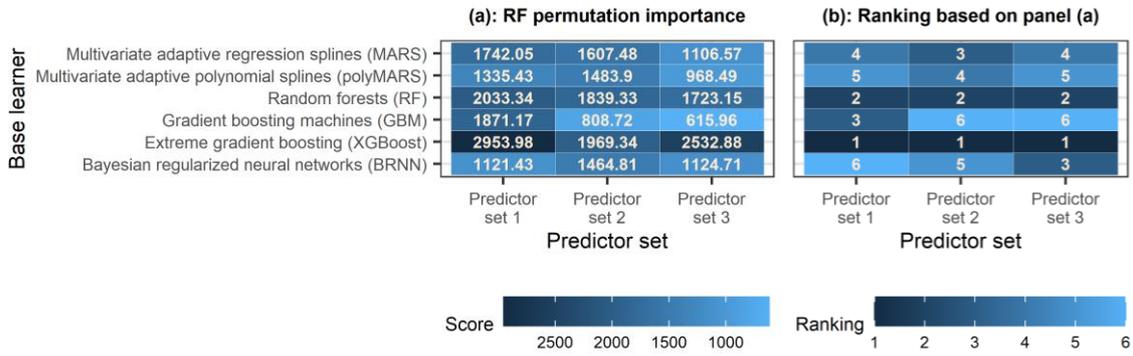

Figure 8. Heatmaps of the: (a) permutation importance score computed using the RF algorithm for the predictions produced by the base learners; and (b) ranking of the base learners based on this score. The darker the colour, the larger the importance.

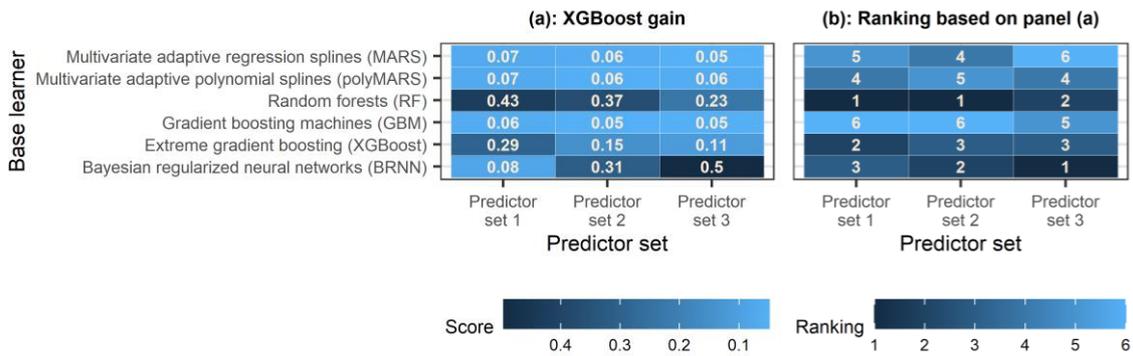

Figure 9. Heatmaps of the: (a) gain score computed using the XGBoost algorithm for the predictions produced by the base learners; and (b) ranking of the base learners based on this score. The darker the colour, the larger the importance.

## 4.4 Importance of predictor variables

In brief, Figure 6 indicates that, for the investigated problem: (a) both the PERSIANN and IMERG datasets offered useful predictors; and (b) the IMERG dataset offered more useful predictors than the PERSIANN dataset, to some extent. Additionally, Figures 10 and 11 present more detailed investigations for predictor variable importance and, more precisely, the permutation importance score computed for the predictor variables using RF, the gain score computed for the same variables using XGBoost and the respective rankings. The comparison based on XGBoost indicates that the IMERG values are more important predictor variables than the PERSIANN values. It also indicates that elevation is more important than the PERSIANN values, and that the PERSIANN and IMERG distances are the least important. This latter finding is also supported by the comparison based on RF, which additionally supports the relatively large importance of elevation as a predictor variable for the problem of interest. On the other hand, the order of the PERSIANN and IMERG values varies a lot in the two comparisons, thereby reinforcing the outcome from Figure 6 that both satellite products offer useful predictors.

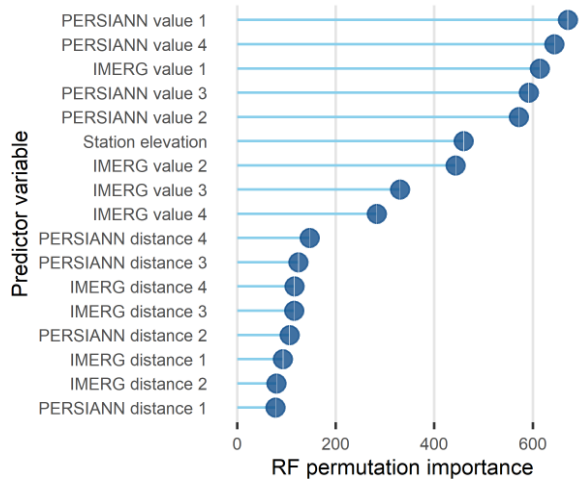

Figure 10. Lollipop of the permutation importance score computed using the RF algorithm for the predictor variables. In the vertical axis, the predictor variables are presented from the most (top) to the least (bottom) important.

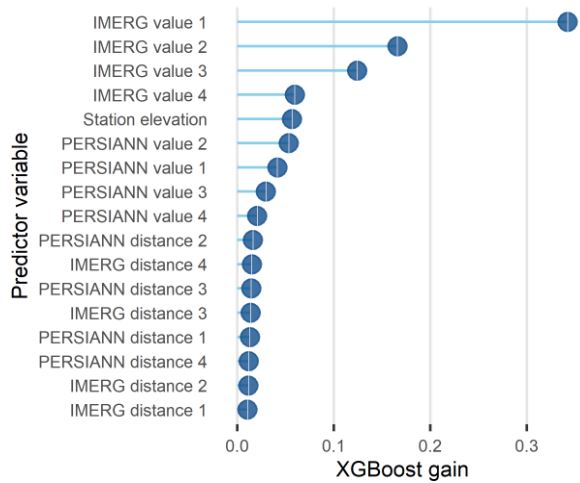

Figure 11. Lollipop of the gain score computed using the XGBoost algorithm for the predictor variables. In the vertical axis, the predictor variables are presented from the most (top) to the least (bottom) important.

## 5. Discussion

This large-scale work is the first one that compares a large number of ensemble learners and combiners in the field of gridded satellite precipitation data correction. In summary, it suggests that stacking a regression algorithm on top of others can lead to large improvements in predictive performance in the context of interest and at the monthly time scale, beating the mean and other simple combiners. Such combiners can be "hard to beat in practice" (Smith and Wallis 2009, Lichtendahl et al. 2013, Winkler 2015, Claeskens et al. 2016, Winkler et al. 2019); therefore, this important finding could not be assumed in advance. In fact, it could only stem from large-scale comparison studies involving both a large number of methods and large datasets.

The improvements in performance were found to be the largest when the combination was made using LR for two of the three predictor sets investigated and polyMARS for the third predictor set. Still, other regression algorithms also performed well as meta-learners, and RF and XGBoost additionally allowed the identification of the most useful base learners (i.e., RF, XGBoost and BRNN) when utilized in explainable machine learning settings. Also notably, the RF and XGBoost algorithms also allowed us to compare the relevance of five categories of predictor variables (i.e., the PERSIANN values, PERSIANN distances, IMERG values, IMERG distances and ground-based station elevation) in solving the problem of interest. Contrary to what applies to the daily time scale (see Papacharalampous et al. 2023b), in the monthly temporal resolution, the utilization of both the PERSIANN and IMERG datasets leads to substantial predictive performance improvements with respect to using the IMERG dataset only.

Although this work already produced a large amount of large-scale results on the use of ensemble learning in the context of improving the accuracy of gridded satellite precipitation products, additional benefits could stem from applying extensions of its methodological framework in the future. Perhaps the most notable among these extensions are the ones referring to the daily time scale (or even to finer time scales), which could additionally benefit from a much larger number of ground-based stations with sufficient record lengths (see, e.g., the number of stations in Tyralis et al. 2023 and Papacharalampous et al. 2023b) and, thus, could lead to comparisons on an even larger scale. Other notable extensions are those referring to probabilistic predictions, and they could comprise machine and statistical learning algorithms, such as those summarized in the reviews by Papacharalampous and Tyralis (2022) and Tyralis and Papacharalampous (2022a). Indeed, it would also be useful to know whether stacking a regression algorithm on top of others can lead to notable performance improvements in the context of assessing the predictive uncertainty in gridded satellite precipitation product correction.

The methodological framework proposed in this work could additionally be extended for investigating whether and how the relative predictive performance of various ensemble learning methods depends on the features of the time series datasets (see the relevant study by Kang et al. 2017). Such investigations would be relevant at various temporal scales, such as the monthly, daily and finer ones, and could eventually lead to the proposal of feature-based ensemble learning methods (see, e.g., those by Montero-Manso et al. 2020, Talagala et al. 2021) for the task of interest. Lastly, it should be noted that, although this study focused on the precipitation domain of earth observation, the framework that it proposed is also applicable to other domains of earth observation. Indeed, the problem of merging gridded satellite and gauge-measured datasets holds a prominent position for earth observation in general.

## 6. Conclusions

To improve the accuracy of gridded satellite precipitation products (i.e., to perform the so-called "bias correction" of these products), machine learning regression algorithms and ground-based measurements are often used in spatial interpolation settings. In such settings, the ground-based measurements are the dependent variables and the gridded satellite data are the predictor variables. In previous studies, individual machine learning algorithms were compared in terms of their skill in performing bias correction of satellite precipitation products. The respective comparisons were conducted either within regional cases or at large spatial scales (i.e., at spatial scales of continental size). While there is value in both these methodological approaches, findings from studies that apply the latter can be considered more useful in the important endeavour of understanding the properties of the algorithms compared. In spite of the accuracy benefits that can stem from ensemble learners (i.e., methods that combine two or more individual machine and statistical learning algorithms to obtain predictive performance improvements), a large number of such learners and their comparison at a large spatial scale were absent from the literature of satellite precipitation product bias correction prior to this work. According to our results, the strategy of stacking a regression algorithm on top of others (which are referred to as "base learners" in the literature) for combining their independent predictions can offer significantly better predictions than the base learners in the context investigated. It also beats the equal-weight combiner, the median combiner and two best learners, one based on the mean squared error and one based on the median squared error.

**Conflicts of interest:** The authors declare no conflict of interest.

**Author contributions:** GP and HT conceptualized and designed the work with input from ND and AD. GP and HT performed the analyses and visualizations, and wrote the first draft, which was commented on and enriched with new text, interpretations and discussions by ND and AD.

**Funding:** This work was conducted in the context of the research project BETTER RAIN (BEnefiTTing from machine lEarning algoRithms and concepts for correcting satellite RAINfall products). This research project was supported by the Hellenic Foundation for Research and Innovation (H.F.R.I.) under the "3rd Call for H.F.R.I. Research Projects to support Post-Doctoral Researchers" (Project Number: 7368).

**Acknowledgements:** The authors are sincerely grateful to the Editor and the Reviewers for their constructive remarks.

**Appendix A      Regression algorithms**

Seven individual regression algorithms were included in the present work for proposing and assessing, in a comparative framework, multiple ensemble learners (see Section 2) for conducting spatial interpolation in the context of blending gridded satellite and gauge-measured precipitation data. In this Appendix, we briefly document these regression algorithms by adapting previous materials by Papacharalampous et al. (2023a and b). The interested reader can find detailed documentations of the same algorithms, for instance, in Hastie et al. (2009), James et al. (2013) and Efron and Hastie (2016). Such documentations are not provided in this work, as they are out of its scope, given that the algorithms are well known in the remote sensing community and that there are automatic software implementations for them. Herein, we used the R programming language (R Core Team 2023) that includes several packages for implementing the selected algorithms. We list the utilized R packages in Appendix B.

The seven individual regression algorithms are the following:

− **Linear regression (LR):** In LR, the dependent variable is a function of a linear weighted sum of the predictors (Hastie et al. 2009, pp 43–55). The weights are estimated by minimizing the mean squared error.

− **Multivariate adaptive regression splines (MARS):** In MARS (Friedman 1991, 1993), the dependent variable is a function of a weighted sum of basis functions. Important parameters are the product degree (i.e., the total number of basis functions) and the knot locations. These parameters are estimated in an automatic way. In our study, we used an additive model and hinge basis functions. We used the default parameters of the R package, as suggested in the respective R package implementation.

- **Multivariate adaptive polynomial splines (poly-MARS):** In poly-MARS (Kooperberg et al. 1997, Stone et al. 1997), the dependent variable is a function of piecewise linear splines within an adaptive regression framework. MARS and poly-MARS differ in the sense that the latter necessitates the existence of linear terms of a predictor variable to be included in the model prior to adding predictor's nonlinear terms, combined with including a univariate basis function in the model prior to including a tensor-product basis function that contains the univariate basis function (Kooperberg 2022). The application was made with the default parameters, as suggested in the respective R package implementation.

- **Random forests (RF):** RF (Breiman 2001) is an ensemble learning algorithm. The ensemble is constructed by decision trees. The construction procedure is based on bootstrap aggregation (also termed as "bagging") with some additional randomization. The latter is based on a random selection of predictors as candidates in the notes of the decision tree. A summary of the benefits of the algorithm can be found in Tyralis et al. (2019b), a study that also comments on the utility of the algorithm in hydrological sciences. The application was made with 500 trees. The remaining parameters of the algorithm were kept equal to their defaults in the respective R package implementation.

- **Gradient boosting machines (GBM):** GBM is an ensemble learning algorithm that trains iteratively new learners on the errors of previously trained learners (Friedman 2001, Mayr et al. 2014, Natekin and Knoll 2013, Tyralis and Papacharalampous 2021). In our case, these learners were decision trees; yet, it is also possible to use other types of learners. The trained algorithm is practically the sum of the trained decision trees. A gradient descent algorithm was used for the optimization. As it is possible to tailor the loss function of GBM to the user's needs, we selected the squared error loss function. We also used 500 trees to be consistent with the implementation of RF. The remaining parameters of GBM were kept equal to their defaults in the respective R package implementation.

- **Extreme gradient boosting (XGBoost):** XGBoost (Chen and Guestrin 2016) is a boosting algorithm that improves over GBM in certain conditions. These conditions are mostly related to data availability. Furthermore, XGBoost is an order of magnitude faster compared to earlier boosting implementations and uses a type of regularization to control overfitting. In our study, we set the number of maximum boosting iterations equal to 500.

The remaining parameters were kept equal to their defaults in the `R` package implementation.

– **Feed-forward neural networks with Bayesian regularization (BRNN):** Artificial neural networks model the dependent variable as a nonlinear function of features that were previously extracted through linear combinations of the predictors (Hastie et al. 2009, p 389). In this work, we applied BRNN (Ripley 1996, pp 143–180, MacKay 1992) that are particularly useful to avoid overfitting. We set the number of neurons equal to 20 and kept the remaining parameters of the algorithm equal to their defaults in the `R` package implementation.

**Appendix B    Statistical software**

We used the `R` programming language (R Core Team 2023) to process the data, to implement and combine the individual regression algorithms (see Appendix A), and to report and visualize the results of the comparison.

For data processing and data visualization, we used the `R` packages `caret` (Kuhn 2023), `data.table` (Dowle and Srinivasan 2023), `elevatr` (Hollister 2022), `ncdf4` (Pierce 2023), `rgdal` (Bivand et al. 2023), `sf` (Pebesma 2018, 2023), `spdep` (Bivand 2023, Bivand and Wong 2018, Bivand et al. 2013) and `tidyverse` (Wickham et al. 2019, Wickham 2023).

For implementing the seven individual regression algorithms, we followed procedures of the `R` packages `brnn` (Rodriguez and Gianola 2022), `earth` (Milborrow 2023), `gbm` (Greenwell et al. 2022), `nnet` (Ripley 2022, Venables and Ripley 2002), `polspline` (Kooperberg 2022), `ranger` (Wright 2023, Wright and Ziegler 2017) and `xgboost` (Chen et al. 2023).

For computing the performance metrics, we used the `R` package `scoringfunctions` (Tyralis and Papacharalampous 2022a, 2022b).

For producing reports, we used the `R` packages `devtools` (Wickham et al. 2022), `knitr` (Xie 2014, 2015, 2023) and `rmarkdown` (Allaire et al. 2022, Xie et al. 2018, 2020).